\documentclass[a4paper,conference]{IEEEtran}
\ifCLASSINFOpdf
   \usepackage[pdftex]{graphicx}
   \usepackage[table,xcdraw]{xcolor}
   \graphicspath{{images/}}
   \DeclareGraphicsExtensions{.pdf,.jpeg,.png,.jpg}
\else
\fi
\usepackage{amsmath}
\ifCLASSOPTIONcompsoc
 \usepackage[caption=false,font=normalsize,labelfont=sf,textfont=sf]{subfig}
\else
 \usepackage[caption=false,font=footnotesize]{subfig}
\fi
\usepackage{url}
\usepackage{xspace}
\makeatletter
\DeclareRobustCommand\onedot{\futurelet\@let@token\@onedot}
\def\@onedot{\ifx\@let@token.\else.\null\fi\xspace}
\def\eg{\emph{e.g}\onedot} 
\def\ie{\emph{i.e}\onedot}

\def\etal{\emph{et al}\onedot}
\makeatother

\usepackage{color}

\usepackage{multirow}

\newcommand{\bx}{\boldsymbol{x}}

\newcommand{\by}{\boldsymbol{y}}

\DeclareMathOperator*{\argmin}{arg\,min}

\usepackage{url}
\usepackage{hyperref}
\begin{document}
\title{Bridging the gap between Natural and Medical Images through Deep Colorization}

\author{
\IEEEauthorblockN{Lia Morra$^1$, Luca Piano$^1$,  Fabrizio Lamberti$^1$,  Tatiana Tommasi$^{1,2}$}
\IEEEauthorblockA{
$^1$ Dipartimento di Automatica e Informatica, 
Politecnico di Torino, 10129, Torino, Italy \\
{$^2$ Istituto Italiano di Tecnologia, Torino, Italy}\\
email: \{name.surname\}@polito.it}
}



\maketitle

\begin{abstract}
Deep learning has thrived by training on large-scale datasets. However, in many applications, as for medical image diagnosis, getting massive amount of data is still prohibitive due to privacy, lack of acquisition homogeneity and annotation cost. In this scenario, transfer learning from natural image collections is a standard practice that attempts to tackle shape, texture and color discrepancies all at once through pretrained model fine-tuning. 
{In this work, we propose to design a dedicated network module that focuses on color adaptation, thus pre-processing the input into a form (RGB) that is closer to the domain the classification backbone was trained on. } We combine learning from scratch of the color module with transfer learning of different classification backbones, obtaining an end-to-end, easy-to-train architecture for diagnostic image recognition on X-ray images. Extensive experiments showed how our approach is particularly efficient in case of data scarcity and provides a new path for further transferring the learned color information across multiple medical datasets.

\end{abstract}


% \fi
\IEEEpeerreviewmaketitle

\section{Introduction}
\label{sec:intro}
In the last years, deep learning has led to major breakthroughs in computer vision, even touching superhuman performance in some task \cite{Silver2016MasteringTG}. One of the crucial ingredients of this success is the availability of large curated image collections on which deep models can be trained. Thanks to the widespread diffusion of RGB cameras, collecting massive amount of images is relatively easy: the expensive part remains getting reliable labels for the image content. This becomes particularly challenging when moving beyond natural images towards specific applications dealing with multispectral information as remote sensing \cite{Bioucas2013remotesensing} or single channel data as in medical diagnosis \cite{Ker2018ieeeaccessmedical}. In particular, the latter needs access to private hospital records and costly manual annotation from experienced doctors. Only very recently some efforts have been done to provide the community with large reference medical datasets \cite{DBLP:journals/corr/abs-1901-07031,rajpurkar2017mura}, which still cover only some specific acquisition modalities,  body parts and pathological markers. 

One way to compensate for the lack of extensive data collections is to transfer knowledge from domains where data are abundant. Although appealing, being able to leverage on photo-based source models for diagnostic tasks means developing approaches to bridge the gap between natural and medical images, and deal with distribution shifts due to texture, shape as well as color. The most standard strategy for this goal is to exploit existing network architectures (\eg ResNet \cite{DBLP:conf/cvpr/HeZRS16}, DenseNet \cite{Huang_2017_CVPR}) pretrained on large datasets like ImageNet \cite{ImagenetJournal}, and simply fine-tune their parameters with a limited amount of medical images \cite{cheplygina2019not}. 
In this way the network maintains a fixed knowledge capacity and the original model's weights are progressively forgotten in favor of those useful to the new domain. A general trade off is adopted between the freedom provided to weight modification and the cardinality of the training data: only larger collections support a full learning of the model parameters and in those cases the pretrained model remains a relevant tool to reduce the training time with respect to learning from scratch. 

\begin{figure*}[th!]
    \centering
    \includegraphics[width=0.9\linewidth]{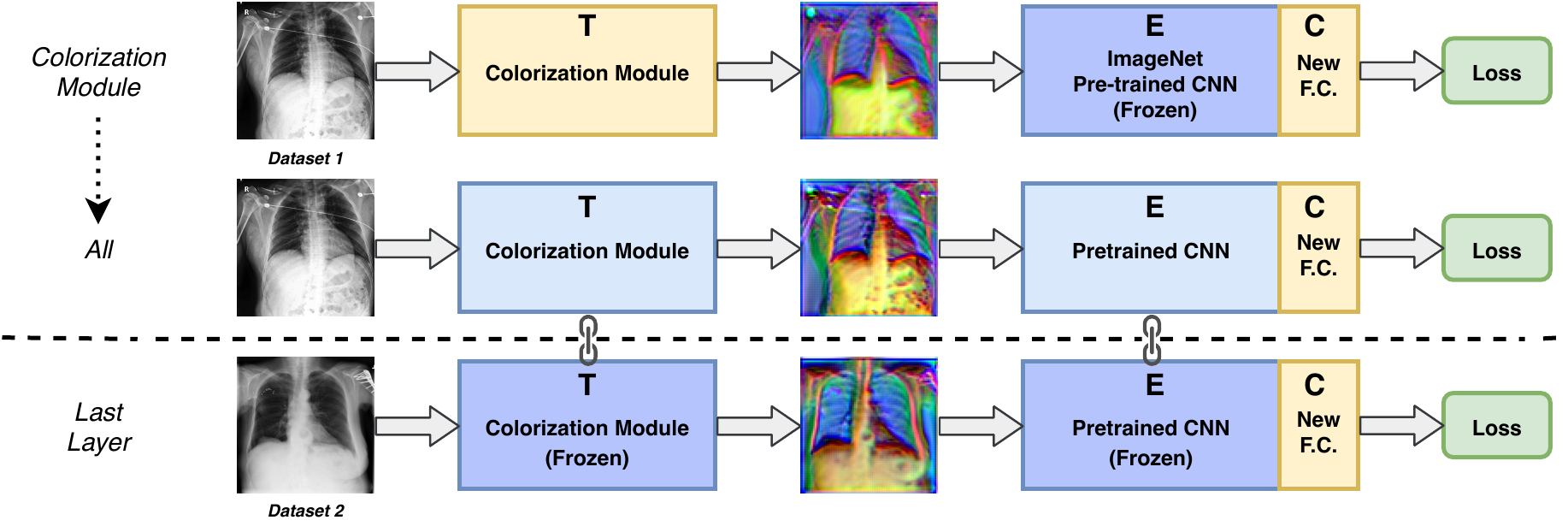}
    \caption{
    A schematic representation of the proposed multi-stage transfer learning pipeline. First, the colorization module $T$ is trained from scratch together with the classifier $C$, while the pre-trained CNN backbone $E$ is kept frozen, to learn the mapping which maximizes classification accuracy. Then, the entire network is fine-tuned to learn useful features for the target task, while simultaneously adjusting the colorization mapping. Finally, the complete trained network is evaluated on a \textit{different} dataset: here we freeze the $\{T,E\}$ modules, and train only the final classification layer $C$.     }
    \label{fig:LearningStrategies}
    \vspace{-2mm}
\end{figure*}

In our work we propose a new transfer strategy that combines the benefit of learning from scratch with that of exploiting pretrained source models, by focusing on color adaptation. The information coded in the three standard color channels for natural images is extremely different from that saved in grayscale X-ray scans, and we show how recomposing this shift provides an essential support to learning with limited annotated data. Specifically, we introduce an efficient colorization module that can be combined with different backbones whose weights are pretrained on ImageNet. The result is an end-to-end, easy-to-train architecture in which the color module is learned from scratch while the backbone can be either kept fixed or finetuned for diagnostic multi-label recognition. Once learned, the colorization module can be further transferred and re-used on medical tasks similar to the primary one (see Fig.~\ref{fig:LearningStrategies} for a schematic overview). 

In the following we first provide a brief summary on previous work about transfer learning in medical imaging, also presenting an overview on existing image colorization approaches (see sec. \ref{sec:related}). Then in section \ref{sec:deco} we describe the details of our multi-stage approach, discussing on how to design the colorization module and presenting three variants for it. Finally section \ref{sec:exper} introduces our experimental setup and section \ref{sec:results} showcases the performance of the proposed method on three state of the art X-ray datasets.  
Finally section \ref{sec:conclusion} summarizes our findings and provides directions for future research.

\section{Related Work}
\label{sec:related}

\subsection{Transfer learning in medical imaging}

Convolutional Neural Networks (CNNs) trained on ImageNet have proven to transfer extremely well to a large variety of medical imaging tasks \cite{cheplygina2019not, tajbakhsh2016convolutional, shin2016deep, samala2018breast,romero2019training, menegola2017knowledge}, to the point that transfer learning from ImageNet has rapidly become a standard practice in the field \cite{litjens2017survey}. However, medical images have distinctive features compared to RGB images: they are usually grayscale, often 12-bit, high resolution images, with a strong textural content. Natural image classification relies predominantly on local contrast, due to changes in illumination conditions, whereas in medical images the intensity values are often informative of the class of tissue class or pathology, and in selected cases may be even standardized. 
Thus, a reasonable question is whether transfer learning from medical images is more effective than transfer learning from RGB (non medical) images or, more in general, which strategies are available to facilitate the transfer from one domain to the other. 

Few head-to-head comparisons are available on \emph{medical vs. non-medical source transfer}, and their results are not always conclusive, also due to different data cardinality, transfer strategies and imaging modalities. A study by Cheplygina \etal \cite{cheplygina2019cats} reviewed twelve papers comparing difference source-target combinations with contrasting results: roughly half of the works indicate that transferring from the non-medical domain outperforms transferring from another medical dataset, with the other half supporting the opposite thesis. Evidence suggests that larger data sets are not necessarily better for pre-training, and that diversity may play an important role, whereas medical datasets tend to be very homogeneous compared to their RGB counterparts \cite{cheplygina2019not}. Pre-training from medical datasets is preferable mostly across the same modality \cite{kim2017modality} and across the same body part \cite{romero2019training, samala2018breast}.
{Previous experiments on Chest X-ray \cite{romero2019training} showed that features learned on medical datasets may be quite specific to a given anatomical site or modality. In fact, while pre-training on another chest X-ray dataset yielded remarkable performance even when fine-tuning on very small datasets, pre-training on images from other anatomical sites (e.g., musculoskeletal X-ray) did not offer any advantage compared to pre-training on ImageNet. }

Fewer studies have compared \emph{transfer learning with training from scratch}. They suggested that fine-tuning pre-trained models is at least as good as training from scratch, and offers substantial advantages for small scale datasets \cite{shin2016deep, cheplygina2019cats}. In recent works on large scale datasets, fine-tuning offered no advantage in terms of performance, but still allowed a remarkable speedup in training \cite{DBLP:journals/corr/Transfusion}. Nonetheless, collecting large scale datasets for all possible modalities and diseases is simply unfeasible, and transfer learning from ImageNet remains one of the prominent strategies for training deep learning models on medical tasks, especially for 2D images.

Alternative transfer solutions are those based on \emph{pseudo-colorization}, but the related literature is rather limited. Pseudo-color images can be generated from CT scans by applying different handcrafted windows/level settings on each channel, in a way similar to how radiologists enhance the contrast of different tissues \cite{shin2016deep}. This technique, however, cannot be transferred to other modalities, such as X-ray, where the intensity values are not standardized. 
Teare \etal designed a genetic algorithm to discover a pseudo-color enhancement scheme for mammography \cite{teare2017malignancy}: each genome encodes three sequences of preprocessing functions along with their parameters (\eg, blurring, contrast enhancement, masking), one for  each color channel, and the preprocessing that maximizes the accuracy of the CNN is selected. This approach, while resulting in a substantial increase in accuracy, is computationally expensive, since it requires training and testing many different networks. To reduce the computational requirements, it is necessary to use a shallower network or a smaller dataset for this step. 

Our work relates to all transfer learning directions described above, but introduces a new colorization-based method for leveraging on non-medical sources. We propose to let the network learn its own optimal color transformation to close the domain gap while solving the medical recognition task of interest. In this way, our approach can take full advantage of all the data available for training at the low cost of updating only a dedicated part of the network, it does not require any strong assumption on which transformations are potentially useful, and can be seamlessly adapted to any type of input modality and any pre-trained CNN architecture.

\subsection{Image colorization}

Image colorization methods generate RGB images starting from a single channel or grayscale one.
Unsurprisingly, in recent years, CNN-based methods have largely outperformed shallow learning models based on hand-crafted features, by leveraging massive amount of natural images from which the color is artificially removed. Indeed, the most common approaches learn a parametric mapping between pairs of corresponding grayscale and RGB images, either explicitly coding the probability distribution of each color \cite{zhang2016colorful,larsson2016learning}, or doing it implicitly through Generative Adversarial Networks \cite{IsolaZZE17,GANcolor}. Of course, those strategies cannot be pursued in the medical domain, where the target color space is unknown. 

Besides the works mainly related to style transfer, colorization techniques have been also investigated for several industrial applications on thermal images \cite{colucci2020automatic} as well as in robotics where single channel depth images are widely used \cite{rgbdcolor, rgbdIEEEaccess}. Of particular interest is the work in \cite{carlucci20172}, in which the authors recently proposed to hallucinate colors from depth through a process guided directly by depth image classification. 
The adopted architecture can take full advantage of the pre-trained ImageNet-based models given that, although significantly different in texture, depth images contain similar objects shapes as their RGB counterparts. 
Inspired by this work, we propose to go beyond it, defining tailored colorization modules for medical images and a  multi-stage transfer procedure involving end-to-end update of the main pretrained backbone.

\section{Transfer learning through colorization}
\label{sec:deco}

\subsection{Learning Strategy}
Let us assume to observe $\mathcal{D}=\{x_i,\by_i\}_{i=1}^N$ labeled data, where $x_i$ is a grayscale medical image and $\by_i \subseteq \{-1,1\}^C$ is its multi-label annotation. Specifically each annotation describes presence $(1)$ or absence $(-1)$
for a set of $C$ pathological clues. Our final goal is to train a classification model able to predict the correct label for each new test sample. To simplify the explanation of the learning procedure, we assign different names to the main components of our deep architecture. We indicate with $E$ the feature encoder, corresponding to the main convolutional backbone of the network.  We use $C$ to specify the final fully connected part of the architecture, whose role is to classify the sample given the features produced by $E$. Standard feature extractors take as input colorful images, thus are not suited to work with grayscale samples. We propose to learn the best colorization function $T: x \rightarrow \bx$ such that the obtained colorful sample $\bx$ can be processed by $E$ and then produce the highest classification performance through $C$. All the parameters of the learning architecture are $\{\theta_E, \theta_C, \theta_T\}$, with the subscript indicating the respective component.

As a first step we leverage on an encoder model $E\big(\bx, \theta^I_E\big)$ pre-trained on ImageNet, keeping it fixed and updating only $T$ and $C$ from scratch. Thus, the objective function is   
\begin{equation}
   \Big \{\theta^M_T, \theta^M_C \Big \}=\argmin_{\theta_T, \theta_C} \sum_{i=1}^N \mathcal{L}\big(C\big(E\big(T(x_i,\theta_T),\theta^I_E\big),\theta_C\big),\by_i\big)~,
   \label{eq:firststep}
\end{equation}
where  $\mathcal{L}$ is the binary cross-entropy loss. In the following we will indicate this learning procedure as \emph{Color Module} (superscript $M$) since the focus is on learning to colorize the original grayscale images. 
Starting from the obtained model, we also define a second learning stage where $T$ and $E$ are fine-tuned, while $C$ is learned from scratch to solve the following optimization problem 
\begin{equation}
\small
    \Big\{ \theta^{A}_T, \theta^{A}_E, \theta^{A}_C \Big \}=  
    \argmin_{\theta_T, \theta_E, \theta_C} \sum_{i=1}^N \mathcal{L}\big(C\big(E\big(T(x_i,\theta_T),\theta_E\big),\theta_C\big),\by_i\big)~,
    \label{eq:secondstep}
\end{equation}
initialized with $\theta_T=\theta^M_T$,  $\theta_E=\theta^I_E$. In the following we indicate this procedure as \emph{All} (superscript $A$).

Till here we were dealing only with a colorization strategy for transfer learning on medical images from an RGB large scale collection as ImageNet. Still, the knowledge acquired in this way can also be shared across different medical image sets. Indeed, we can transfer the learned component $T$, when facing  new labeled observations $\mathcal{B}=\{x_i,\by_i\}_{i=1}^{N_B}$, with  $\mathcal{B} \neq \mathcal{D}$. Here the difference among the datasets can be of various nature: similar body parts and task but different annotation procedure, or even completely different body parts and labeling. By transferring $T$ and keeping it fixed together with $E$, the problem reduces to solving 
\begin{equation}
   \Big \{\theta^{L}_C \Big \}=\argmin_{\theta_C} \sum_{i=1}^N \mathcal{L}\big(C\big(E\big(T(x_i,\theta_T),\theta_E\big),\theta_C\big),\by_i\big)~.
   \label{eq:thirdstep}
\end{equation}
We use the tag \emph{Last Layer} (superscript $L$) to indicate this strategy, which can have different variants, depending on which %modules %
$\{T,E\}$ are used as starting point, 
with the two options  $\{\theta_T^M, \theta_E^I\}$ or $\{\theta_T^{A}, \theta_E^{A}\}$. 
Alternatively, the $\{T,E\}$ components could be transferred and further fine-tuned while learning $C$ from scratch, with the same optimization problem in eq. (\ref{eq:secondstep}). 

At test time, the prediction on a new image will be obtained as $\tilde{\by}=\big(C\big(E\big(T(x,\theta^*_T),\theta^*_E\big),\theta^*_C\big)$ with the superscript $^*$ indicating the parameters obtained at the end of the chosen learning strategy.

\begin{figure*}[htb]
    \centering
    \includegraphics[width=0.9\linewidth]{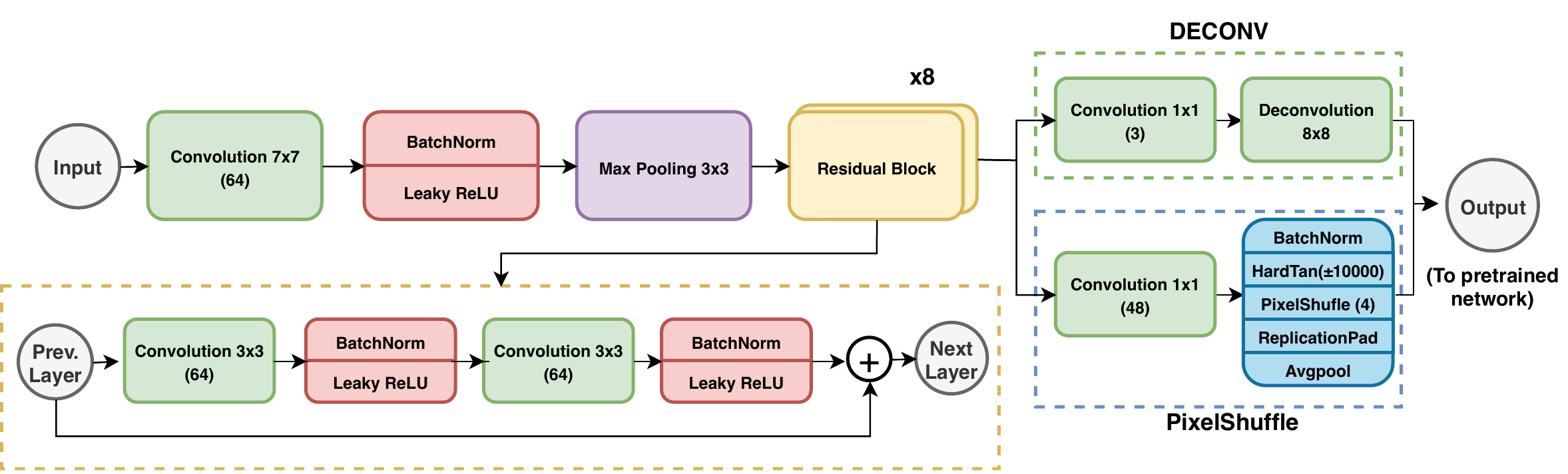}
    \caption{Overview of the DECONV and Pixel Shuffle colorization modules: their share the main structure with a difference only in the final upsampling layer.
    Details of the residual block are shown in the dashed, bottom box. }
    \label{fig:Modules}
    %\vspace{-2mm}
\end{figure*}

\subsection{Colorization Module}
The main challenge at the basis of the described learning approach concerns how to design the colorization module $T$. Building over \cite{carlucci20172}, we consider two architectures that use a convolution and pooling layer to transform the $x=\{1\times320\times320\}$ grayscale image into a $\bx=\{64\times80\times80\}$ tensor, which is then given as input to a sequence of 8 residual blocks, as shown in Fig. \ref{fig:Modules}.
The two architectures differ in the final up-sampling part: we indicate as 
\emph{DECONV} the case using a Transpose Convolution as in \cite{carlucci20172}, and with \emph{Pixel Shuffle}
our modification based on the homonym technique originally developed for super-resolution applications \cite{pixelshuffle_Shi_2016_CVPR}. This technique allows to reduce the checkerboard artifacts introduced by up-sampling.

\begin{figure}[tb]
    \centering
    \includegraphics[width=0.25\textwidth]{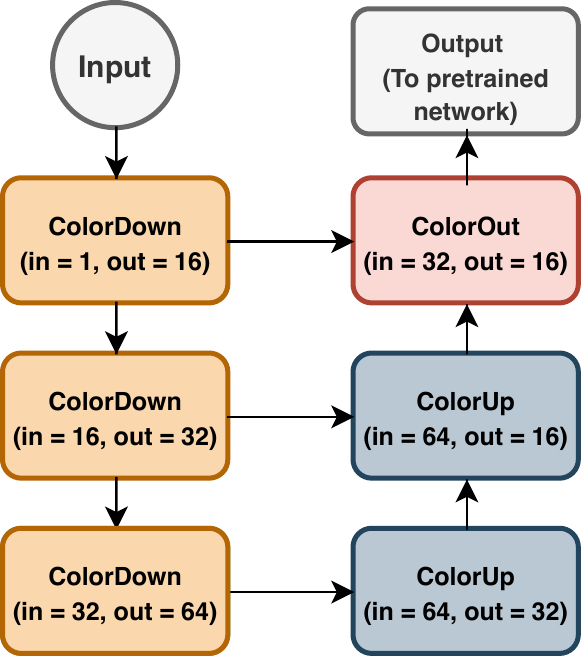}
    \caption{ColorU architecture. The input image passes through 3 down-sampling blocks (\emph{ColorDown}), 2 up-sampling blocks (\emph{ColorUp}) and a final up-sampling and colorization block block (\emph{ColorOut}). For each block, the number of input and output channels is reported.}
    \label{fig:ColorU}

\end{figure}

We also defined a third variant of the colorization module\footnote{{All our trained colorization modules are available at  \url{https://gitlab.com/grains2/DeepMedicalColorization}}}, indicated in the following as \emph{ColorU} (Fig.\ref{fig:ColorU}). It takes inspiration from the U-Net architecture \cite{unet_miccai2015,Billaut2018ColorUNetAC} but presents different blocks.
The input image passes through 3 down-sampling blocks (\emph{ColorDown}), 2 up-sampling blocks (\emph{ColorUp}) and another final up-sampling block (\emph{ColorOut}) that generates the colorful output image. All these blocks use strided convolutions instead of spatial pooling functions, allowing the models to learn how to up/down-sample rather than using fixed methods, as proposed in   \cite{springenberg2014striving}. Each ColorDown block is composed by two convolutional layers (conv with kernels $3 \times 3$ and $4 \times 4$) followed by a BatchNorm Layer. Each ColorUp block is composed by a Transpose Convolutional layer ($4 \times 4$), followed by two conv layers and a BatchNorm layer. The ColorOut model has a similar structure to the ColorUp, but without the final BatchNorm layer. All conv layers have a Leaky ReLU activation, applied before BatchNorm, except for the final conv layer which outputs the colorized image. 
ColorU is much more efficient  than DECONV and Pixel Shuffle, with a ratio of about $1/3$ in terms of needed learning parameters.

\section{Experiments}
\label{sec:exper}
\subsection{Datasets}
\subsubsection{Chest X-ray}
We experimented on two large chest X-ray datasets, CheXpert \cite{DBLP:journals/corr/abs-1901-07031} and ChestX-ray14 \cite{wang2017chest}. 

The CheXpert training set contains 224,316 chest radiographs from 65,240 patients, whereas ChestX-ray14 contains 112,000 images from 32,717 patients. Both datasets are labeled for the presence or absence of 14 independent observations starting from radiological reports using two different semi-automatic labellers. We focus here on five pathological cues: Atelectasis, Cardiomegaly, Consolidation, Edema and Pleural Effusion, which are common to both datasets and for which prior results were available in literature \cite{DBLP:journals/corr/abs-1901-07031}. 

CheXpert ground truth takes into account uncertainty in the reporting and labeling process by assigning each observation one of three values: positive, negative and uncertain. Different policies were previously compared in \cite{DBLP:journals/corr/abs-1901-07031} to handle uncertain labels during training: converting uncertain to positive  (U-Ones) or negative values (U-Zeros), or ignoring uncertain samples (U-Ignore).  For each observation, we applied the best policy based on previous results: Atelectasis (U-Ones), Cardiomegaly (U-Zeros), Consolidation (U-Zeros), Edema (U-Ones), and Pleural Effusion (U-Ones) \cite{DBLP:journals/corr/abs-1901-07031}. {We did not employ any specific training strategy to account for data imbalance during training.}

\subsubsection{MURA}
MURA  \cite{rajpurkar2017mura} comprises 40,561 multi-view musculoskeletal radiographs from about 12,000 patients, representing various body parts as either normal or abnormal. For our analysis we considered only one anatomical part, the shoulder, corresponding to 8,379 images (2,694 patients).  

\subsection{Experimental setup}

We performed experiments on two baseline networks, with the $E$ module of our architecture corresponding to either ResNet18 or DenseNet121.

When working on CheXpert and ChestX-ray14 the output layer is the combination of five binary heads ($C=5$), one for each observation. For MURA the task reduces to a simple binary classification problem with $C=1$.

Standard data augmentation is applied during training: non-square images are randomly cropped and rescaled to the network input size ($320 \times 320$), then random rotation  {($-10^{\circ}$ -- $+10^{\circ}$}) and zoom (0\% -- 10\%) are applied with 75\% probability. All the images were normalized with the mean and standard deviation calculated over CheXpert. 
All the networks were trained until convergence using the SGD optimizer with the One Cycle Policy \cite{DBLP:journals/corr/abs-1708-07120}. The Learning Rate (LR) finder was used to determine the maximum learning rate for each network \cite{smith2017cyclical}. Weights were saved every 4800 iterations and we used the checkpoint with the highest validation performance. 

\subsection{Evaluation}
Performance was assessed using the Area under the ROC curve (AUC). For multi-label datasets (CheXpert and ChestX-ray14), the AUC was calculated for each observation, and then 
the mean AUC was used to summarize the performance. For each configuration, the training was repeated three times with different random initialization. Paired t-test was used to compare different transfer learning procedures. We controlled for multiple hypothesis testing using the Benjamini--Hochberg procedure  \cite{benjamini1995controlling}; an adjusted p-value $<0.05$ indicated statistical significance.

\section{Results}
\label{sec:results}

\subsection{Transfer learning through colorization}

In this section we compare the different transfer learning procedures illustrated in section \ref{sec:deco} with the standard approach, \ie without the colorization module. Specifically, the baseline model takes as input a three channel image obtained by copying the single channel sample, and is constituted by the encoder $E$, followed by final fully connected component $C$. 
The baseline can be trained by freezing the weights $\theta_E^I$ (denoted in the following as \textit{Baseline}) or fine-tuning all the layers (denoted in the following as \textit{Baseline All}). For the colorization module $T$, we adopted the multi-stage transfer procedure described in Section  \ref{sec:deco}: first we train $T$ from scratch (\emph{Color Module}), and then both $T$ and $E$ are fine-tuned, while $C$ is learned from scratch (\emph{All}). We report the performance after each phase. 

Average AUC values for the ResNet18 and DenseNet121 architectures are shown in Table \ref{tab:CX_MU_E}. Training the color module $T$ with fixed $E$ improves classification performance  over the Baseline by roughly 7\%: results are statistically significant for both ResNet18 (AUC=78.4 vs. 83.4, $p$=0.037) and DenseNet121 (AUC=78.6 vs. 84.3, $p$=0.043). These results confirm that color is indeed an important component of the domain shift from RGB to medical images. The obtained accuracy is still lower with respect to Baseline All, which however requests the update of $10^7$ network parameters against the $10^5$ needed with the Color Module solutions. A further fine-tuning of the whole network with the All strategy gets us back to the Baseline All results (ResNet18: AUC=89.6 vs. 88.9, $p$= 0.078; DenseNet121: AUC=89.8 vs. 89.6, $p$=0.142).

\emph{Regarding the specific choice of $E$}, both DenseNet121 and ResNet18 achieve similar performance. DenseNet121 seems to offer a small advantage over ResNet18 in conjuction with PixelShuffle (AUC=89.6 vs. 88.9, $p$=0.10), but the baseline models are comparable (AUC=89.8 vs. 89.6, $p$=0.4). These findings are in agreement with previous studies discussing how networks characterized by different depth behave similarly on medical datasets \cite{DBLP:journals/corr/Transfusion}. 
This is likely due to the limited number of classes (compared to the thousand classes of ImageNet) and relatively homogeneous image content.

\begin{table}[tb]
\centering
\caption{Average AUC for different transfer learning strategies on CheXpert. }
\label{tab:CX_MU_E}
\resizebox{0.48\textwidth}{!}{%
\begin{tabular}{|c|c|c|}
\hline

\textbf{Colorization Module} & \textbf{Learning Strategy} & \textbf{Mean AUC} \\ \hline
\rowcolor[HTML]{EFEFEF} 
\multicolumn{3}{|c|}{\cellcolor[HTML]{EFEFEF}\textbf{Backbone: ResNet18 }} \\ \hline
\rowcolor[HTML]{DAE8FC} 
 -  & Baseline & 78.4 $\pm$ 0.5 \\ \hline
DECONV & Color Module & 84.0 $\pm$ 0.3 \\ \hline
PixelShuffle & Color Module & 83.4 $\pm$ 0.4 \\ \hline
ColorU & Color Module & 83.9 $\pm$ 0.8 \\ \hline
\rowcolor[HTML]{DAE8FC} 
 - & Baseline All & 89.6 $\pm$ 0.2 \\ \hline
DECONV & All & 88.9 $\pm$ 0.3 \\ \hline
PixelShuffle & All & 88.9 $\pm$ 0.1 \\ \hline
ColorU & All & 89.3 $\pm$ 0.3 \\ \hline
\rowcolor[HTML]{EFEFEF} 
\multicolumn{3}{|c|}{\cellcolor[HTML]{EFEFEF}\textbf{Backbone: DenseNet121}} \\ \hline
\rowcolor[HTML]{DAE8FC} 
 - & Baseline & 78.6 $\pm$ 0.2 \\ \hline
DECONV & Color Module & 83.5 $\pm$ 0.3 \\ \hline
PixelShuffle & Color Module & 84.3 $\pm$ 0.8 \\ \hline
ColorU & Color Module & 83.9 $\pm$ 0.5 \\ \hline
\rowcolor[HTML]{DAE8FC} 
 - & Baseline All & 89.8 $\pm$ 0.2 \\ \hline
DECONV & All & 89.2 $\pm$ 0.2   \\ \hline
PixelShuffle & All & 89.6 $\pm$ 0.1  \\ \hline
ColorU & All & 89.7 $\pm$ 0.3  \\ \hline
\end{tabular}%
}
\end{table}

\begin{table}[tb]
    
       \caption{Results for individual observations on the CheXpert dataset with DenseNet121. The results from \cite{DBLP:journals/corr/abs-1901-07031} are reported for completeness although obtained with an ensemble of 30 checkpoints.}
    \label{tab:ResultsByObservation}
    \centering
    \resizebox{0.48 \textwidth}{!}{%
        \begin{tabular}{|c|c|c|c|c|c|}
            \hline
            \textbf{Learning }                     &  \textbf{Atelectasis} & \textbf{Cardiomegaly} & \textbf{Consolidation} & \textbf{Edema} & \textbf{Pleural }  \\ 
           \textbf{ strategy}    &  \textbf{} & &  & & \textbf{ Effusion}  \\ \hline
            DenseNet121 \cite{DBLP:journals/corr/abs-1901-07031}  & 85.8
            & 84.0              &  93.2      & 94.1 & 93.4 \\ \hline\hline

            Baseline All   & \textbf{85.4 $\pm$ 0.5}
            & 82.7  $\pm$ 0.6                &  \textbf{94.9$\pm$ 1.0}  & 93.2  $\pm$ 0.3 & 89.8   $\pm$ 0.2     \\ \hline
            PixelShuffle All   & 84.7 $\pm$ 1.7
            & \textbf{85.0  $\pm$ 1.8}                 &  93.8     $\pm$ 0.8   & \textbf{93.8   $\pm$ 0.9} & \textbf{91.0  $\pm$ 0.6}     \\ \hline

        \end{tabular}%
    }
\vspace{-2mm}
\end{table}
\begin{figure*}[bt]
    \centering
    \includegraphics[width=0.9\linewidth]{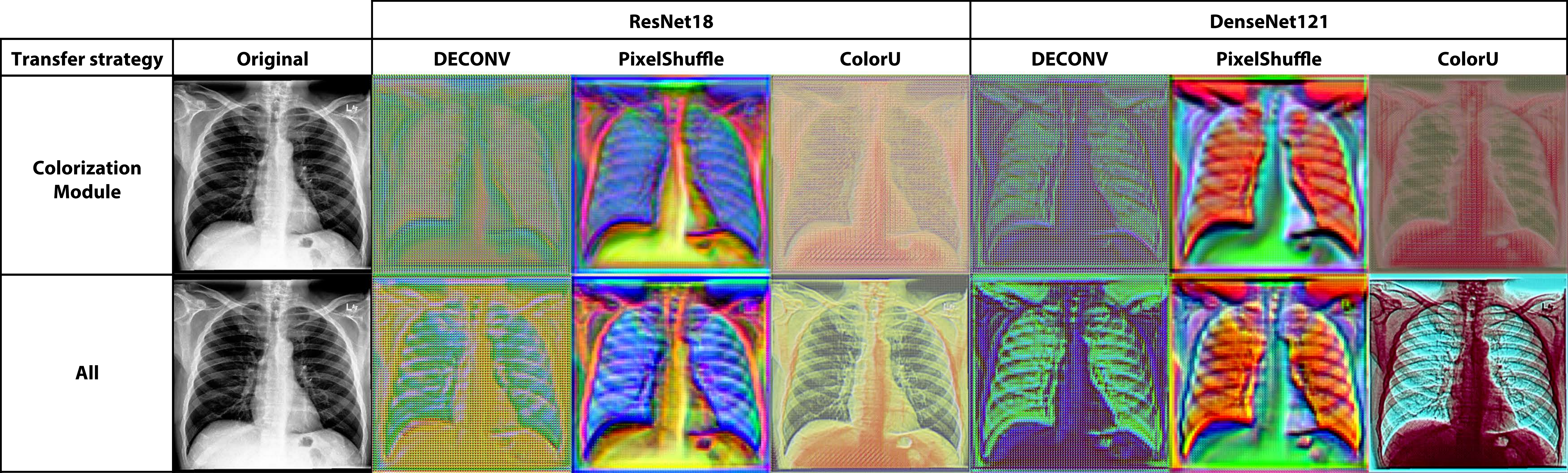}
    \caption{Output of different colorization modules optimized for the ResNet18 and DenseNet121 architectures. In the top row, only the colorization module $C$ is trained, freezing the encoder $E$. The bottom row shows results after fine-tuning all the layers end-to-end.}
    \label{fig:paper_images}

\end{figure*}

\emph{Regarding the specific choice of $T$}, we can claim that all the considered variants offer similar advantages in terms of performance.
We found a small, albeit not significant, advantage of ColorU over PixelShuffle for ResNet18 (AUC=83.9 vs. 83.4, $p$=0.132), but not for DenseNet121 ($p$=0.60); PixelShuffle is always equal or better than DECONV.
Overall, differences are numerically small and suggest that colorization is stable across a wide range of architectures and encoders, including both very deep and more shallow networks.

\emph{A different point of view on the results is offered by the colorful images obtained as a byproduct of the learning process.}
As shown in Figure \ref{fig:paper_images}, they depend both on the specific used module $T$ and on whether $E$ is fixed or fine-tuned. 
The output of the colorization module is suggestive of the network learning process: anatomical features become more prominent after the whole network is fine-tuned, suggesting that the pre-trained encoder is somewhat unable to identify important image features and more information on shape and texture need to be adapted besides the overall color. We underline that the goal of our work is not that of producing human-readable images, rather that of letting the  network assign the most useful colors to get the maximum recognition accuracy. Whether and to what extent the colorization highlights areas that are relevant to the final diagnosis is an interesting avenue for future work.

Finally, \emph{a per-class analysis} on the Pixel Shuffle All DenseNet121 results reveals that adding the colorization module increases the performance on Cardiomegaly ($p$=0.128) and  Edema ($p$=0.024). 
On the other hand, Baseline All outperforms PixelShuffle All for Consolidation ($p$=0.07). Differences are less decisive for Atelectacsis ($p$=0.40) and Pleural Effusion ($p$=0.37) (see Table \ref{tab:ResultsByObservation}). 
Only for this evaluation we are endorsing statistical significance for $p<0.15$ to take into consideration the {higher} variability among different runs in every class.
The obtained results suggest that different learning strategies may converge to different feature representations, with comparable overall performance, but complementary strengths and weaknesses.

\subsection{Effect of training set size}
\begin{figure}[tb]
    \centering
    \subfloat{\includegraphics[width=0.48\columnwidth]{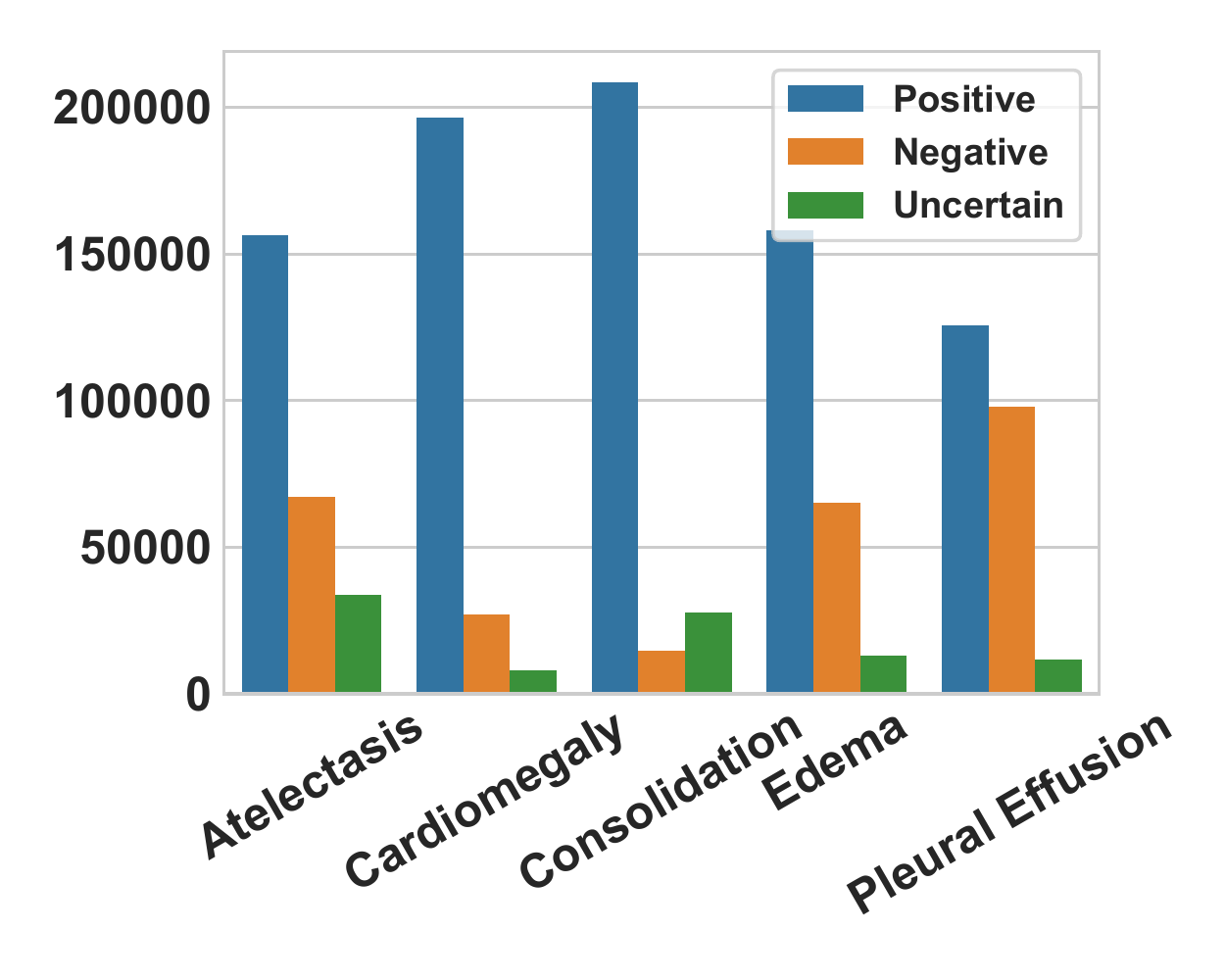}
    }
    \hfil
    \centering
    \subfloat{\includegraphics[width=0.48\columnwidth]{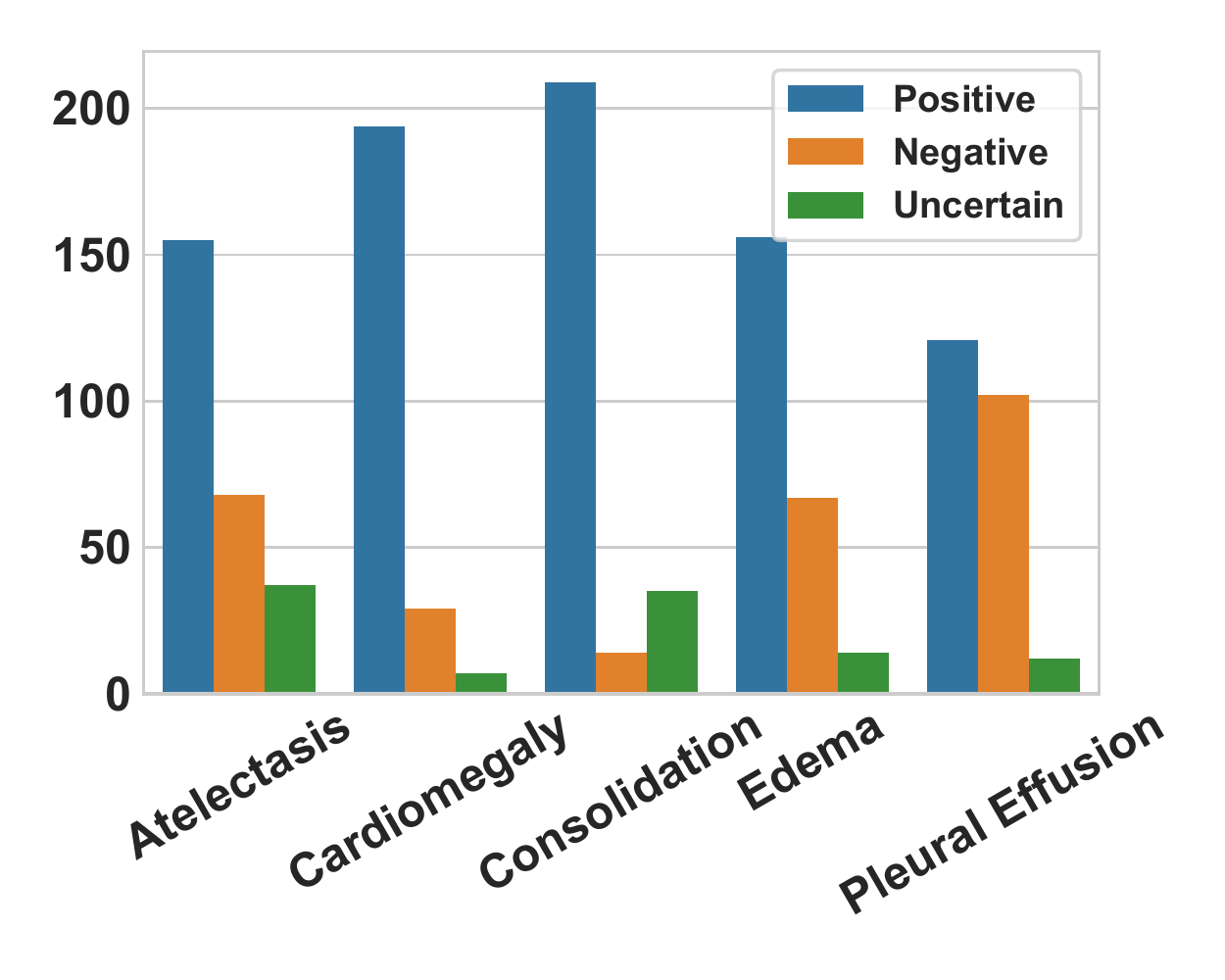}
   }
    \caption{Number of positive, negative and uncertain labels per observation in the entire CheXpert dataset (left) and in the 0.1\% subset (right). The proportion of positive, negative and uncertain cases is the same across all dataset sizes. } 
    \label{fig:distribution}
  
\end{figure}

The optimal transfer policy naturally depends on the size of the training set.
Since the proposed colorization strategy allows to perform transfer learning on a budget of learning parameters with respect to whole network fine-tuning, we expect its advantage to become more visible in case of limited amounts of training data. 
CheXPert, while significantly smaller than ImageNet, is by far larger than most medical datasets. To study the effect of colorization in the small and very small data regimes, we conducted additional experiments on subsets ranging from 10\% (22,431 images) to 0.1\% (224) of the original CheXPert size. This range is representative of small dataset (5,000-10,000 images) to few-shot learning scenarios. Indeed, the distribution of positive, negative and uncertain labels varies widely among classes (see Figure \ref{fig:distribution}) and, in the 0.1\% scenario, the number of positive examples can be as low as 23. 
For each sampling size, three random datasets were selected; we report the average and standard deviation of the AUC across the three datasets. All experiments were conducted with the ResNet18 backbone and the PixelShuffle colorization module. 

\begin{figure*}[tb]
    \centering
    \subfloat{\includegraphics[width=0.4\linewidth]{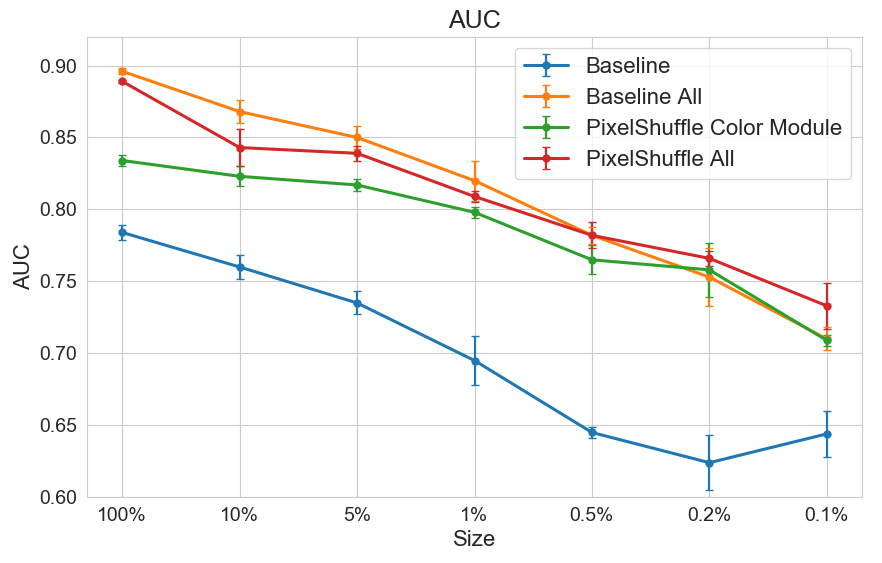}
    \label{fig_first_case}}
    \hfil
    \centering
    \subfloat{\includegraphics[width=0.4\linewidth]{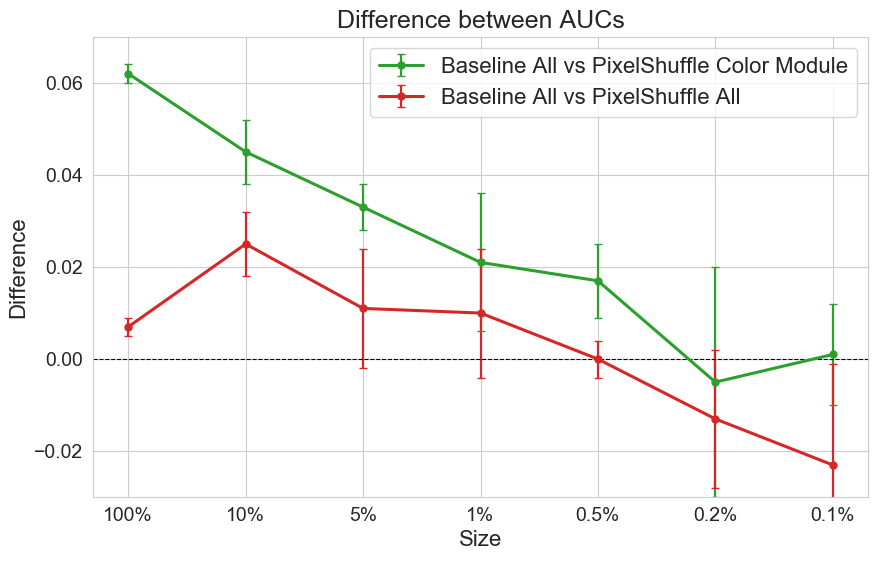}
    \label{fig_second_case}}
    \caption{Performance of different transfer strategies according to the dataset size. (a) Average AUC for four learning strategies: without colorization module (Baseline) and with PixelShuffle Color module. Results are reported both with the baseline is kept frozen (blue and green lines) as well as when fine-tuning the entire network (yellow and red lines). (b) Difference between the PixelShuffle architecture and the fine-tuned baseline network. When the all the architecture is fine-tuned (red line), PixelShuffle outperforms the baseline architecture in the limited data scenario. Training only the colorization module (green line) is less effective, but the difference progressively reduces for smaller datasets. }
    \label{fig_sim}
    \vspace{-3mm}
\end{figure*}

We compare the baseline and the colorization-enhanced network in two settings: when the encoder $E$ is frozen (PixelShuffle Color Module vs Baseline), and when $E$ is fine-tuned (Pixel Shuffle All vs Baseline All).
As shown in Figure \ref{fig_sim}, regardless of the dataset size, the Baseline with frozen weights is outperformed by all the other configurations, indicating that even a very small dataset is sufficient to learn better feature representation for this task. Adding a colorization module consistently improves results in all data regimes, and the gap in performance is much higher in the small data regime (400 -- 2,000 images). 

PixelShuffle is equal (Color Module) or better (All) than  Baseline All in the very small data regime (400 -- 2,000 images).  The proposed multi-stage transfer procedure (PixelShuffle All) outperforms training only the colorization module (PixelShuffle Color Module) across all dataset sizes, thus training $T$ as a first step provides a more effective starting point for learning novel features useful for the new task. In the small data scenario, the network has limited ability to learn new features; instead, the cheap colorization module can be effectively trained from scratch. On larger datasets, the advantage of the colorization module is reduced and the limited increase in parameter count may explain the lower performance. It should be noticed that all these experiments were conducted on a rather shallow backbone (ResNet18), and that the advantage of the proposed technique could be even higher on deeper models, where the chance of overfitting will be proportionally higher.

\begin{table}[tb!]
\centering
\caption{Average AUC for different transfer learning strategies across datasets. All the experiments are performed with ResNet18 backbone.
}
\label{tab:MURAC14}
\resizebox{0.48\textwidth}{!}{%
\begin{tabular}{|c|c|c|c|}
\hline

\textbf{Backbone $E$} & \textbf{Colorization Module $T$} & \textbf{Learning strategy} & \textbf{Mean AUC} \\ \hline
\rowcolor[HTML]{EFEFEF} 
\multicolumn{4}{|c|}{\cellcolor[HTML]{EFEFEF}\textbf{Mura}} \\ \hline
\rowcolor[HTML]{DAE8FC} 
($\theta_E^{I}$) ImageNet & - & Baseline & 73.0 $\pm$ 0.11 \\ \hline
($\theta_E^{I}$) ImageNet & PixelShuffle ($\theta_T^{M}$ CheXpert) & Last Layer & 67.1 $\pm$ 1.2 \\ \hline

\rowcolor[HTML]{EFEFEF} 
\multicolumn{4}{|c|}{\cellcolor[HTML]{EFEFEF}\textbf{ChestX-ray14}} \\ \hline
\rowcolor[HTML]{DAE8FC} 
($\theta_E^{I}$) ImageNet & - & Baseline & 66.9 $\pm$ 0.1 \\ \hline
($\theta_E^{I}$) ImageNet & PixelShuffle ($\theta_T^{M}$ CheXpert) & Last layer & 68.3 $\pm$ 1.4 \\ \hline
($\theta_E^{I}$) ImageNet & PixelShuffle ($\theta_T$ Scratch) & Color Module & 72.9 $\pm$ 0.1 \\ \hline
($\theta_T^{A}$) CheXpert & PixelShuffle ({$\theta_T^{A}$} CheXpert) & Last layer & 77.3 $\pm$ 0.5 \\ \hline
\rowcolor[HTML]{DAE8FC} 
($\theta_E^{I}$) ImageNet & - & Baseline All & 79.9 $\pm$ 0.06 \\ \hline
($\theta_E^{I}$) ImageNet & PixelShuffle ($\theta_T^{M}$ CheXpert) & All & 79.4 $\pm$ 0.06 \\ \hline
($\theta_E^{I}$) ImageNet & PixelShuffle ($\theta_T$ Scratch) & All & 79.2 $\pm$ 0.3 \\ \hline
({$\theta_T^{A}$}) CheXpert & PixelShuffle ({$\theta_T^{A}$} CheXpert) & All & 81.0 $\pm$ 0.1 \\ \hline

\end{tabular}%
}
%\vspace{-2mm}
\end{table}

\subsection{Transferability across datasets}

We evaluated transferability of the colorization module $T$ on two datasets, MURA and ChestX-ray14, representing two distinct scenarios: same modality/different body parts (MURA), and same modality/same body part (ChestX-ray14). Both MURA and ChestX-ray14 are smaller than CheXpert, which can therefore act as an intermediate step between RGB images and the medical domain. Multiple factors need to be taken into account in designing the transfer strategy: which encoder $E$ is used as starting point  ($\{\theta_T^M, \theta_E^I\}$ or $\{\theta_T^{A}, \theta_E^{A}\}$), whether $T$ and $E$ are frozen or further fine-tuned on the target dataset. As in previous experiments, the baseline model is defined by the encoder $E$ and by the final layer $C$; the latter is trained from scratch in all configurations. 

On MURA, when starting from the initial configuration $\{\theta_T^M, \theta_E^I\}$, the performance degrades with respect to the baseline (AUC=67.1 vs. 73.0, $p$=0.040). Differently, on ChestX-ray14, the performance largely increases when the colorization module is transferred ($\{\theta_T^M, \theta_E^I\}$, AUC=68.3 vs. 66.9, $p$= 0.005), as well as when both colorization module and backbone are transferred ($\{\theta_T^{A}, \theta_E^{A}\}$, AUC=77.3 vs. 66.9, $p$=0.017).  These results strongly suggest that the colorization module transfers well to similar datasets, but not across different body parts. Recent literature have also discussed how transferring from ImageNet is better than transferring from a medical source collection capturing a different body part than the target one \cite{romero2019training}. {This behavior likely arises from the lack of variety in the CheXpert dataset, which induces the network to learn very specific features and/or colorization patterns.} Hence, we did not further insist on MURA, moving our focus towards more experiments on ChestX-ray14.

Transferring both the colorization module and backbone from CheXpert ($\{\theta_T^{A}, \theta_E^{A}\}$, Pixel Shuffle Last Layer) achieves slightly inferior results with respect to the Baseline All on ChestX-ray14 (AUC=77.3 vs. 79.9, $p$=0.016). 
The best results are obtained when the network is pre-trained on CheXpert and then fine-tuned on ChestX-ray14 ($\{\theta_T^{A}, \theta_E^{A}\}$, Pixel Shuffle All, AUC=81.0 vs. 79.9, $p$=0.008). 
We remark that, even though the task between CheXpert and ChestX-ray14 is the same, there is still a considerable domain shift 
between the two datasets, both due to the acquisition and pixel intensity distribution, as well as to the labelling process. 
Nonetheless, pre-training on CheXpert, also through the colorization module, can offer a significant boost in performance, which we expect would be even higher across more similar datasets or in case of extremely reduced data availability.

\section{Conclusions}
\label{sec:conclusion}
This paper presented a novel strategy for transfer learning from the non-medical to the medical domain. It leverages a deep colorization module to learn an optimal color mapping for a given pre-trained convolutional neural network. We propose a multi-stage transfer learning procedure where the lightweight colorization module is first trained from scratch, keeping the backbone frozen, followed by a refinement stage where both modules are fine-tuned towards the target task. 

Our approach produces a significant gain in performance when the feature encoder is frozen with two different backbones and over two datasets. These results indicate that learning how to colorize medical images already helps to reduce the gap with RGB data from the non-medical domain with a minimal computational effort. 
Further fine-tuning the backbone is helpful to fully adapt to differences in shape and texture, besides color. This is confirmed by visual analysis of the hallucinated color images.

%The experiments in the small and very small data regime highlighted how the best transfer learning procedure strongly depends on the number of training images
% and confirms the power of our colorization multi-stage transfer procedure, which opens a new path for transferring learned color information to similar medical datasets (\ie containing the same body part).

{The experiments in the small and very small data regime highglighted how our colorization-based multi-stage transfer procedure is particularly effective when the target dataset is small, which is often the case in many biomedical applications. When a large dataset is available, as in the case of CheXpert, it opens a new path for transferring learned color information to similar medical datasets (\ie containing the same body part).} 

Future work will investigate the effectiveness and generality of our approach on different medical imaging datasets, different pre-trained backbones, and different tasks beyond classification. The proposed colorization modules can be easily extended to manage multiple inputs, thus offer a promising solution for leveraging on pre-existing source models even when the target samples are 3D and multi-channel (\eg magnetic resonance imaging).


%\section*{Acknowledgment}




\bibliographystyle{IEEEtran}
\bibliography{IEEEabrv,root}


% \end{thebibliography}

\end{document}